\documentclass[conference]{IEEEtran}
\usepackage{standalone}
\usepackage[textsize=tiny]{todonotes}
\usepackage{tabularx}
\usepackage{pgfplots, pgfplotstable}
\usepackage{textcomp}
\usepackage{makecell}
\usepackage{booktabs}
\usepackage{listings}
\usepackage{lipsum}
\usepackage{multirow}
\usepackage{color}
\usepackage[T1]{fontenc}

\makeatletter
\newcommand\BeraMonottfamily{%
  \def\fvm@Scale{0.85}
  \fontfamily{fvm}\selectfont
}
\makeatother

\usepackage{xcolor}
\usepackage{array}
\usepackage{wasysym}
\usepackage[flushleft]{threeparttable}

\usepackage{hyperref}

\begin{document}

\title{PHE-SICH-CT-IDS: A Benchmark CT Image Dataset for Evaluation Semantic Segmentation, Object Detection and Radiomic Feature Extraction of Perihematomal Edema in Spontaneous Intracerebral Hemorrhage}

\author{
 \IEEEauthorblockN{
 Deguo Ma\IEEEauthorrefmark{1}, 
 Chen Li \IEEEauthorrefmark{1}, 
 Lin Qiao\IEEEauthorrefmark{2}, 
 Tianming Du\IEEEauthorrefmark{1}, 
 Dechao Tang\IEEEauthorrefmark{1}, 
 Zhiyu Ma\IEEEauthorrefmark{1}, 
 Liyu Shi\IEEEauthorrefmark{1},
 }

 \IEEEauthorblockN{
 Marcin Grzegorzek\IEEEauthorrefmark{3} and 
 Hongzan Sun\IEEEauthorrefmark{2} }
 
 \IEEEauthorblockA{\IEEEauthorrefmark{1} Microscopic Image and Medical Image Analysis Group, College of Medicine and Biological Information Engineering, \\ 
 Northeastern University, Shenyang, China}
 \IEEEauthorblockA{\IEEEauthorrefmark{2} Shengjing Hospital, China Medical University, Shenyang, China}
 \IEEEauthorblockA{\IEEEauthorrefmark{3} Institute of Medical Informatics, University of Luebeck, Luebeck, Germany}
}

\maketitle

\thispagestyle{plain}
\pagestyle{plain}

\begin{abstract}

Background and objective: Intracerebral hemorrhage is one of the diseases with the highest mortality and poorest prognosis worldwide. Spontaneous intracerebral hemorrhage (SICH) typically presents acutely, prompt and expedited radiological examination is crucial for diagnosis, localization, and quantification of the hemorrhage. Early detection and accurate segmentation of perihematomal edema (PHE) play a critical role in guiding appropriate clinical intervention and enhancing patient prognosis. However, the progress and assessment of computer-aided diagnostic methods for PHE segmentation and detection face challenges due to the scarcity of publicly accessible brain CT image datasets.

\noindent Methods: This study establishes a publicly available CT dataset named PHE-SICH-CT-IDS for perihematomal edema in spontaneous intracerebral hemorrhage. The dataset comprises 120 brain CT scans and 7,022 CT images, along with corresponding medical information of the patients. To demonstrate its effectiveness, classical algorithms for semantic segmentation, object detection, and radiomic feature
extraction are evaluated. The experimental results confirm the suitability of PHE-SICH-CT-IDS for assessing the performance of segmentation, detection and radiomic feature extraction methods.

\noindent Results: This study conducts numerous experiments using machine learning and deep learning methods, demonstrating the differences in various segmentation and detection methods on the PHE-SICH-CT-IDS. The highest precision achieved in semantic segmentation is 76.31$\%$, while object detection attains a maximum precision of 97.62$\%$. The experimental results on radiomic feature extraction and analysis prove the suitability of PHE-SICH-CT-IDS for evaluating image features and highlight the predictive value of these features for the prognosis of SICH patients.

\noindent Conclusion: To the best of our knowledge, this is the first publicly available dataset for PHE in SICH, comprising various data formats suitable for applications across diverse medical scenarios. We believe that PHE-SICH-CT-IDS will allure researchers to explore novel algorithms, providing valuable support for clinicians and patients in the clinical setting. PHE-SICH-CT-IDS is freely published for non-commercial purpose at: \url{https://figshare.com/articles/dataset/PHE-SICH-CT-IDS/23957937}.

\end{abstract}

\bstctlcite{IEEEexample:BSTcontrol}

\IEEEpeerreviewmaketitle


\section{Introduction}
\label{s:Introduction}

\subsection{Research background and motivation}

Intracerebral hemorrhage (ICH) is a type of cerebrovascular accident resulting from bleeding within the brain tissue, leading to the accumulation of blood.
This disorder is known to be caused by various factors, among which hypertension is the most common, accounting for about 65\% of spontaneous cases. Other causes include amyloid angiopathy, brain tumors, aneurysms, arteriovenous malformations, cerebral cavernous malformations, and arteriovenous fistulae. These underlying causes may lead to the rupture of blood vessels, resulting in the formation of a hematoma within the brain parenchyma~\cite{2012Intracerebral}. 
ICH remains a significant cause of morbidity and mortality worldwide, with an estimated incidence of 2.8 million cases annually. ICH accounts for approximately 10\% to 15\% of all strokes in the USA, Europe, and Australia, and up to 20\% to 30\% of strokes in Asia~\cite{2019Heart, 2010Advances}. The incidence rate of ICH is 24.6 per 100,000 person-years, making it a considerable public health concern~\cite{2015Emergency}. The consequences of ICH can be severe, including long-term disability and death. The financial burden of ICH is attributed in part to its high mortality, with up to 40\% of patients succumbing to the condition within 30 days of onset, often after prolonged stays in the intensive care unit~\cite{2010Incidence}. Moreover, the incidence of ICH is projected to rise due to the increasing use of anticoagulation, antiplatelet drugs, and an aging population.

Spontaneous intracerebral hemorrhage (SICH) is a frequent subtype of ICH, frequently occurring in the basal ganglia. Patients with SICH may experience early re-bleeding. SICH typically presents acutely, and timely imaging is crucial for accurate diagnosis, determination of the location and volume of bleeding.

The cranial computed tomography (CT) imaging is preferred in the stroke unit due to its prevalence, non-invasiveness, and affordability in nearly all hospitals, while offering good quality visual information on human organs.
In CT images, the hemorrhage appears as a hyper-intense bright region with sharp contrast against its surroundings and the perihematomal edema (PHE) as a low-density area around the hemorrhage. 
Numerous preclinical and clinical studies have demonstrated that the kinetics and peak volume of PHE have been shown to cause secondary brain injury (SBI) after SICH and is associated with a poor prognosis. Thus, PHE has been considered a promising therapeutic target for SICH.
Furthermore, the CT images of PHE also provide several radiomic feature parameters that can predict hematoma expansion (HE). 
Patients with SICH may experience HE in the early stage, which can increase the mortality rate of the patient. HE is a critical determinant of disease progression and poor prognosis~\cite{2011Defining}.
Practical scoring schemas have been developed based on these parameters and clinical criteria to predict HE accurately. 
However, localization of PHE regions in CT images is extremely challenging due to significant overlap in CT values between HE and other brain tissues, such as cerebrospinal fluid and microvascular diseases. For experienced radiologists, detecting intracerebral hemorrhage using pixel labeling requires an average of 10 minutes per CT scan, while PHE labeling requires at least twice as much time. 
The present diagnosis process involves the examine of CT images by an expert radiologist to determine the presence of ICH and identify its type and location. 
However, the diagnosis is dependent on the availability of subspecialty-trained neuroradiologists, leading to potential time inefficiencies and inaccuracies in remote areas where access to specialized care is limited. 
The current research status reveals several critical issues in the field: medical professionals face challenges in accurately diagnosing and localizing Spontaneous Intracerebral Hemorrhage (SICH) due to the subjective nature of results, excessive workload, and extended working hours. Furthermore, medical doctors lack radiomic features parameters to predict HE. Therefore, there is a pressing need to address these relevant issues more effectively.

With the development of medical image processing technology, the primary goal is to achieve high accuracy and good performance in computer-assisted diagnosis. To accomplish this objective, several deep learning-based methods have been proposed and explored~\cite{ronneberger2015unet}. Specifically, medical image segmentation and detection are crucial for achieving this goal. The rapid development of object detection methods has enabled the efficient and rapid identification of hemorrhage and PHE locations. In addition, the results of image segmentation methods can provide important reference for clinical doctors in predicting the risk of HE and predicting patient's cycle survival rate. Overall, the medical image processing technology holds great potential for improving the accuracy and efficiency of medical diagnosis for SICH.

In this paper, a benchmark CT image dataset for evaluation semantic segmentation, object detection, and radiomic feature extraction of perihematomal edema in spontaneous intracerebral hemorrhage is introduced, namely PHE-SICH-CT-IDS, which is constructed 120 CT scans of patients with SICH. PHE-SICH-CT-IDS contains 3,511 CT images of SICH occurring in the basal ganglia region, with associated labels for the surrounding edematous zone around the hematoma. Additionally, PHE-SICH-CT-IDS provides SICH detection labels for the edematous zone and  radiomic features of the edematous zone, which are valuable for clinical research and diagnosis. The evaluation results are obtained by using different traditional machine learning and new deep learning methods for segmentation and detection on the images of the CT dataset. Dataset is available at the URL:\url{https://figshare.com/articles/dataset/PHE-SICH-CT-IDS/23957937}.

The main contributions of this paper are as follows:  
\begin{itemize}
  \item [1)] 
  The first open-source CT dataset of spontaneous intracerebral hemorrhage with perihematomal edema in the basal ganglia region (PHE-SICH-CT-IDS) is developed and released. 
  \item [2)]
  PHE-SICH-CT-IDS provides segmentation and detection capabilities for edematous areas, as well as radiomic features of the edematous areas around intracerebral hemorrhage, which can be used by medical researchers to study the correlation between PHE and HE within a short period of time. 
  \item [3)]
  The validation of the related edematous area segmentation and detection methods proposed by individuals highlights the distinguishability of PHE-SICH-CT-IDS for commonly used machine learning and deep learning methods.
\end{itemize}

\subsection{Related work}

This study analyzes and compares existing public CT datasets on ICH, and explores in-depth the currently known research results. Additionally, it highlights the limitations of the relevant datasets currently available.

Two commonly used public datasets for ICH detection are the CQ500 dataset~\cite{chilamkurthy2018development, chilamkurthy2018deep} and the RSNA Intracranial Hemorrhage Detection Challenge dataset(RSNA ICHD dataset). The CQ500 dataset, originating from the Centre for Advanced Research in Imaging, Neurosciences, and Genomics in New Delhi, India, encompasses a diverse range of CT scanners with slices per rotation varying from 16 to 128. CQ500 includes 491 head CT scans with each of intraparenchymal, subdural, extradural, and subarachnoid hemorrhages, as well as calvarial fractures.  
RSNA ICHD dataset, which was released by the Radiological Society of North America (RSNA) in collaboration with members of the American Society of Neuroradiology and MD.ai for the RSNA Challenge on Kaggle, and contains over 25,000 CT scans. 

For ICH segmentation, the PhysioNet ICH dataset contains 82 CT scans. Among these, 36 scans are from patients diagnosed with ICH by radiologists, including types such as intraventricular, intraparenchymal, subarachnoid, epidural, and subdural~\cite{MD2020Intracranial, hssayeni2020computed}. Each CT scan has around 30 slices with a 5 mm slice thickness.The number of images in this dataset is small and the variety is insufficient.

In addition to ICH detection and segmentation, many approaches have been proposed for PHE segmentation~\cite{chen2022defining, kuang2022uncertainty}. However, a lack of public or private datasets with PHE masks has prevented the validation of many of these approaches. Others have been validated on private datasets with different characteristics, such as the region of onset in the brain and the type of ICH diagnosed. Given these differences, it is not possible to objectively compare different approaches. Therefore, a dataset is required to benchmark and extend the work on PHE segmentation in ICH.

\begin{table*}[!htbp]
\centering
\setlength{\abovecaptionskip}{0pt}
\setlength{\belowcaptionskip}{10pt}
\caption{Recent datasets for segmentation and detection of intracerebral
hemorrhage.}
\label{tab:aStrangeTable}
    \begin{tabular}{@{}cccccc@{}}
    \toprule
    Year  & Name    & Reference  & Category   & Amount & modality\\
    \midrule
    2018  & CQ500     &	~\cite{chilamkurthy2018development, chilamkurthy2018deep}      & Detection, Classification	&  491 scans    & CT\\
    2019   & RSNA Intracranial Hemorrhage Detection Challenge dataset      & ~\cite{flanders2020construction}     & Detection	& 25,000 scans    & CT\\
    2020        & PhysioNet ICH dataset    & ~\cite{MD2020Intracranial, hssayeni2020computed}      & Segmentation  & 36 scans    & CT\\
    2022        & ATLAS V2.0   & ~\cite{liew2022large}      & Segmentation  & 955 scans    & MRI\\
    \bottomrule
    \end{tabular}
\end{table*}

\begin{figure*}[!htbp]
    \centering
    \includegraphics[width=0.9\textwidth]{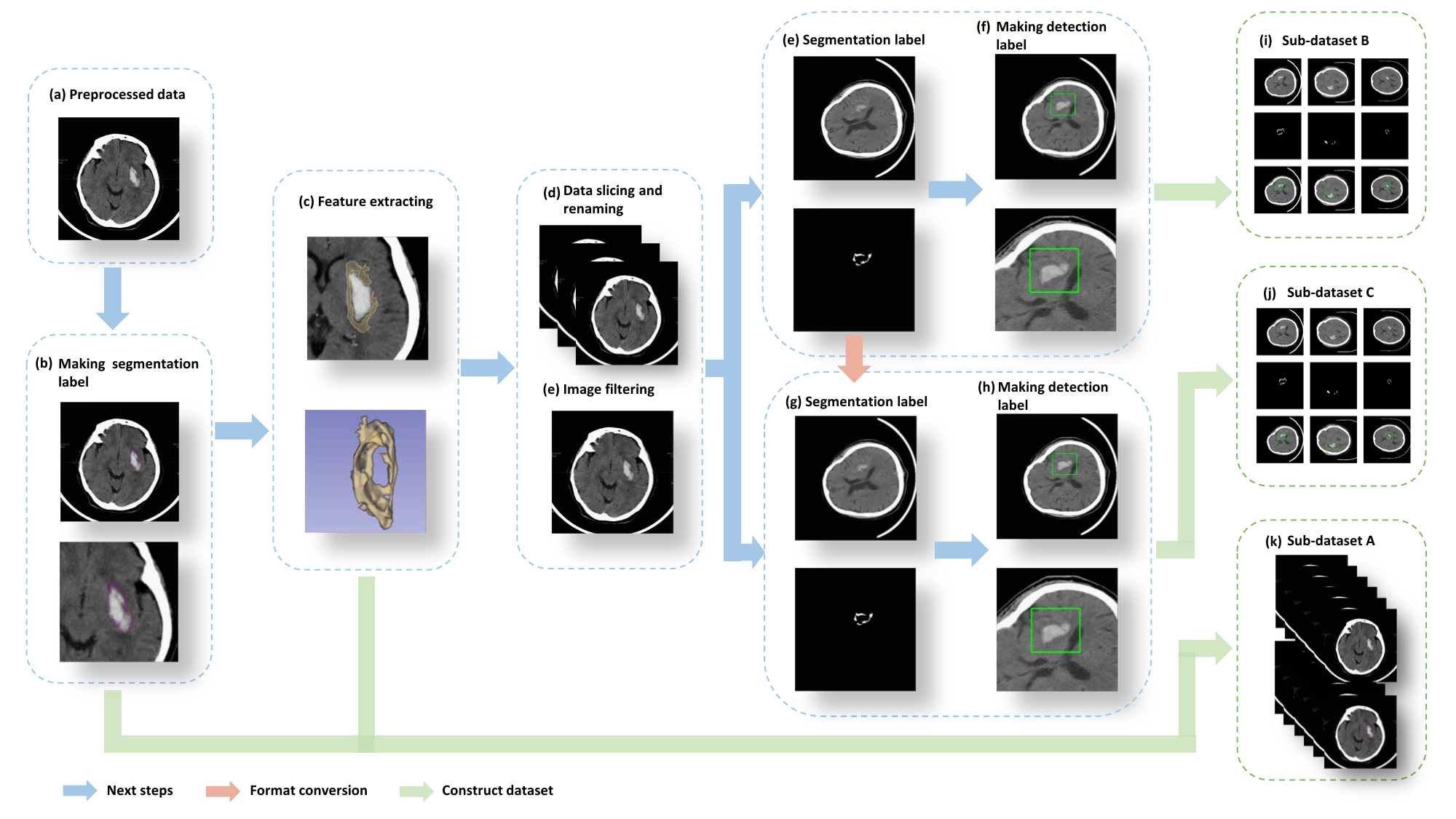}
    \caption{: Data preparation workflow of PHE-SICH-CT-IDS.}
    \label{fig:Dataset workflow}
\end{figure*}

\subsection{Structure of this paper}

In this section, the background and motivation for the dataset preparation are presented along with a review of related research papers. Detailed methods for preparing and evaluating the dataset, encompassing information about each individual element. Section~\ref{s:Result} presents the results of the dataset evaluation, followed by a discussion of these results in section~\ref{s:Discussion}. Finally, a summary of the findings and potential future work are presented.

\section{Materials and Methods}
\label{s:Materials and Methods}

\subsection{Dataset preparation}
\label{ss:Dataset preparation}
PHE-SICH-CT-IDS includes 7,022 CT images, consisting of 3,511 CT slice images and 3,511 ground truth images of PHE. Moreover, 3,511 object detection labels in XML format are provided. Additionally, PHE-SICH-CT-IDS contains 120 CT scans and corresponding PHE labels in NIFTI format for eligible patients. Radiomic features of the PHE are also available for these 120 patients. The dataset includes medical information for these patients such as presence of the subsequent hematoma expansion, gender, age, diagnostic haemorrhagic area and previous medical history (e.g., hypertension and diabetes). In response to the different functions of the dataset and the diverse requirements of experiments, the dataset is divided into three sub-datasets, with specific information provided in Table~\ref{s:DatasetTable}. The subsequent section provides an introduction to the specifics of the applied datasets.


The complete process, starting from data solicitation and concluding with the compilation of the final dataset, is depicted in Figure~\ref{fig:Dataset workflow}.

\begin{figure*}[!htbp]
    \centering
    \includegraphics[width=0.9\textwidth]{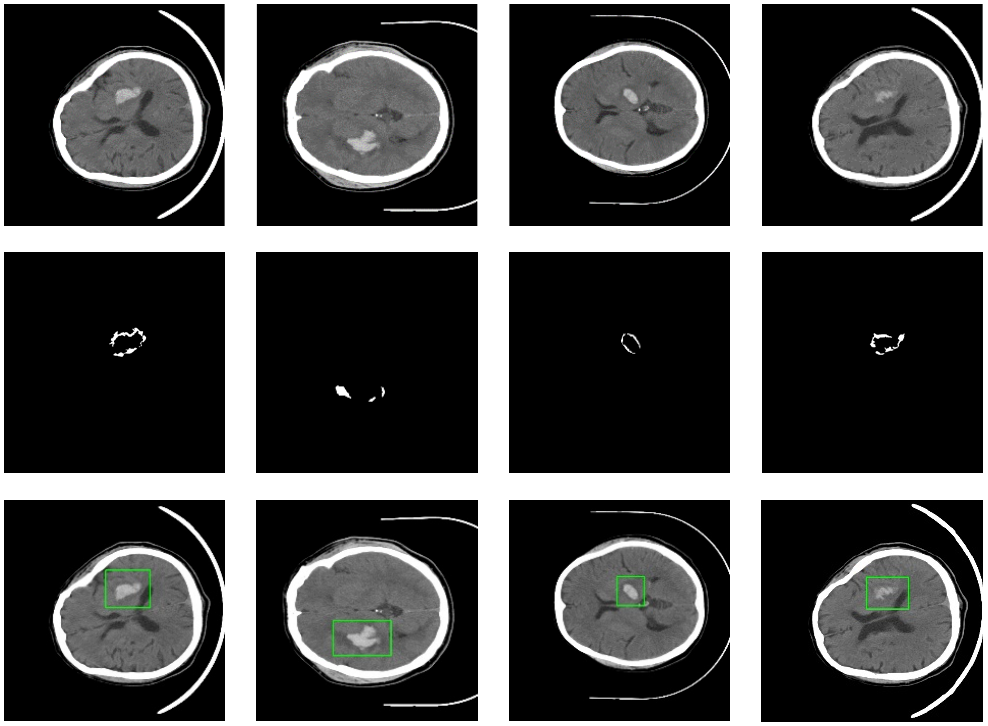}
    \caption{The labels of SICH-CT-IDS. The first row represents the original CT images, the second row represents the segmentation labels, and the third row represents the detection labels.}
    \label{fig:lableExample}
\end{figure*}

PHE-SICH-CT-IDS:

\begin{enumerate}

\item Data source:

\par
The head CT scan data of SICH was collected between December 2020 and April 2023 from Shengjing Hospital of China Medical University. The CT scans were rigorously screened by three senior radiologists with 15, 17, and 20 years of experience in cranial CT interpretation, according to strict criteria(See section~\ref{sss:Data selection criteria} for details). The CT scans were performed using a PHILIPS Brilliance iCT scanner with the following parameters: tube voltage 120 kV, tube current 300 mA, matrix $512 \times 512$, slice thickness 5.00 mm, and slice spacing 5.00 mm. Each CT scan consisted of an average of approximately 35 slices.

\item Data selection criteria: 
\label{sss:Data selection criteria}
\par Inclusion criteria:
    \begin{itemize}
      \item [1)] 
      Diagnosis of spontaneous intracranial hemorrhage.       
      \item [2)]
      Hemorrhage occurring in the basal ganglia region of the brain. 
      \item [3)]
      Time from SICH symptom onset to hospital admission and scanning within 12 hours.
      \item [4)]
      Age over 18 years.
      \item [5)]
      Patients who did not undergo surgical treatment after admission and reexamined their head CT within 72 hours.
    \end{itemize}

\par Exclusion criteria:
    \begin{itemize}
      \item [1)] 
      Poor image quality.       
      \item [2)]
      History of neurosurgical procedures.
      \item [3)]
      Coagulation disorders or a history of anticoagulant use.
      \item [4)]
      Secondary ICH or bleeding into the ventricles
    \end{itemize}

\item Perihematomal edema segmentation labels: 
\par During the data collection process, if the selection criteria are met, the open-source software LIFEx v5.10 is used for PHE segmentation. Using the cerebral window settings (Level=40, Width=120), radiologists manually outline the PHE range on each CT image layer. The outlining is based on the density of the PHE. Notably, this study exclusively focuses on outlining areas with low-density PHE. The specific segmentation labels are shown in Figure~\ref{fig:lableExample}.

\item Object detection labels: 
\par Annotations for object detection are created using the Labelme annotation tool and saved in PASCAL VOC format as XML files with the same name as the corresponding images. The annotated regions include not only the area of intracranial hemorrhage but also the surrounding PHE. The detection labels in the dataset are presented in Figure~\ref{fig:lableExample}.

\item Radiomic features:
\par Based on the segmented PHE regions, the open-source software LIFEx v5.1 is used for radiomic feature extraction, which yield a total of seven texture features. These include Homogeneity, Energy, Contrast, Correlation, Entropylog2, Entropylog10, and Dissimilarity, all derived from the gray-level co-occurrence matrix (GLCM).

\item Dataset preparation workflow:
    \begin{itemize}
      \item [1)] 
      Selecting data that meets the criteria.       
      \item [2)]
      Retaining patient-related medical information.
      \item [3)]
      Creating segmentation labels. (Figure~\ref{fig:Dataset workflow}-(b))
      \item [4)]
      Extracting radiomic features. (Figure~\ref{fig:Dataset workflow}-(c))
      \item [5)]
      Converting original data and segmentation label data in NIFIT format to PNG format slices with adjusted window level and width. (Figure~\ref{fig:Dataset workflow}-(d))
      \item [6)]
      Creating object detection labels. (Figure~\ref{fig:Dataset workflow}-(f)(h))
    \end{itemize}

\item Sub-dataset and image format:
    \begin{itemize}
      \item [1)] 
      Sub-dataset A: NIFIT format.       
      \item [2)]
      Sub-dataset B: PNG format and $512 \times 512$ pixels.
      \item [3)]
      Sub-dataset C: JPG format and $512 \times 512$ pixels.
    \end{itemize}
\end{enumerate}

\begin{table*}[!htbp]
\centering
\setlength{\abovecaptionskip}{0pt}
\setlength{\belowcaptionskip}{10pt}
\caption{Dataset content of PHE-SICH-CT-IDS.}

\label{s:DatasetTable}
    \begin{tabular}{@{}ccccc@{}}
    \toprule
    Subdataset name  & Format   & Amount  & Function & Others\\
    \midrule
    Subdataset A   & NIFIT   & 120 scans      & Segmentation and Feature extraction	& Radiomic features Medical information\\
    Subdataset B   & PNG    & 120 scans / 7,022 images     & Segmentation and Detection	& Medical information\\
    Subdataset C   & JPG    & 120 scans / 7,022 images      & Segmentation and Detection  & Medical information\\
    \bottomrule
    \end{tabular}
\end{table*}

\subsection{Dataset description}
\label{ss:Dataset description}

    \subsubsection{Normal CT Image}
    \label{sss:Normal CT Images}
    A brain CT image of a healthy individual reveals normal brain structure and tissue characteristics. The cortex and white matter regions exhibit normal density and contrast, with well-defined gray-white matter boundaries. The ventricular system appears in a normal size and shape, and the sulcal patterns and cerebral gyri are intact and clearly visible. No signs of abnormal hemorrhage, masses, edema, or other pathological changes were observed. Overall, the image depicts a healthy brain with normal anatomical structure and tissue integrity. Some examples are shown in Figure~\ref{fig:exampleSICH}. 
    
    \subsubsection{CT image of SICH}
    \label{sss:CT images of SICH}
\begin{figure*}[!htbp]
        \centering
        \includegraphics[width=0.9\textwidth]{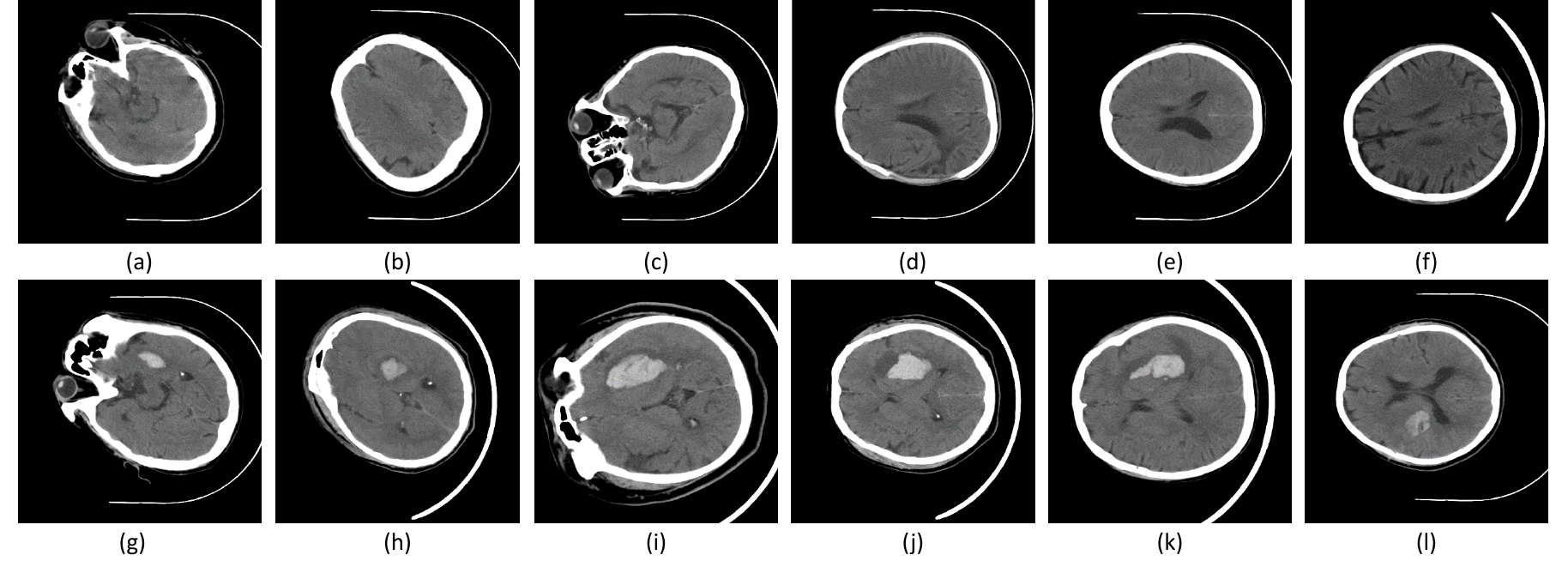}
        \caption{Example of CT images: (a) (b) (c) (d) (e) (f) Normal CT images without SICH, (g) (h) (i) (j) (k) (l) CT images of SICH.}
        \label{fig:exampleSICH}
    \end{figure*}

    During CT examinations, ICH is characterized by uniformly high-density images in an elongated or circular shape, while PHE manifests as low-density images surrounding the site of hemorrhage. Examples of the related content can be observed in Figure~\ref{fig:exampleSICH}.
    
\subsection{Methods of segmentation}
\label{ss:Methods of segmentation}
Five classical machine learning methods, namely $k$-means, MRF, Otsu and Watershed, are employed for segmenting PHE-SICH-CT-IDS. Additionally, this study employs four distinct deep learning techniques for the segmentation of PHE-SICH-CT-IDS. These methods, classified as either classical or novel, encompass U-Net, SegNet, SwinUNet, and TransUNet.

\subsubsection{Segmentation machine learning methods}
\label{sss:Segmentation machine learning methods}

$k$-means clustering algorithm is a well-known partitioning clustering and segmentation method. It is widely used in image segmentation~\cite{2015Image}.
$k$-means clustering for extracting shape features from images characterized by low contrast and backgrounds exhibiting multiple levels of variation brings forth several advantages. These encompass clear contours, efficient algorithmic processing speed, and optimal utilization of memory resources. Therefore, it is an effective algorithm for grayscale image segmentation.

Markov Random Field (MRF) is a model described using an undirected graph~\cite{2004Unsupervised}. It consists of a set of nodes, where each node corresponds to a single variable or a group of variables. The links between nodes are undirected. In the context of image segmentation, it can be viewed as an image clustering problem. It involves grouping pixels with similar properties into the same class.

Otsu's method, also known as the Otsu's thresholding algorithm, was proposed in 1979 for determining the threshold value for image binarization~\cite{liu2009otsu}. In terms of the principle of Otsu's method, it is also known as the maximum between-class variance method. This is because the threshold obtained through Otsu's method results in the maximum between-class variance when applied to image binarization, separating the foreground and background. It is considered the optimal algorithm for threshold selection in image segmentation due to its simplicity, independence from image brightness and contrast, and wide application in digital image processing.

Watershed segmentation is a mathematical morphology-based method rooted in topological theory~\cite{shojaii2005automatic}. Its fundamental concept is to treat an image as a topographic landscape in geodesy, where the grayscale value of each pixel represents its elevation. Each local minimum and its influence region are referred to as a catchment basin, while the boundaries between catchment basins form the watershed.

\subsubsection{Deep learning methods}
\label{sss:Deep learning methods}

U-Net is primarily used for addressing challenges in biomedical image analysis and has gained popularity in the field of medical image segmentation~\cite{ronneberger2015unet}. It employs classical design methods such as downsampling, upsampling, and skip connections. Due to its U-shaped network structure, it is referred to as U-Net. The left half represents the feature extraction part, while the right half represents the upsampling part, also known as the encoder-decoder structure. The left side can be viewed as an encoder, and the right side can be viewed as a decoder. Additionally, the network employs skip connections by connecting the upsampling results with the outputs of the encoder submodules that have the same resolution. These connections serve as inputs for the subsequent submodules in the decoder.

UNet++ incorporates a series of short connections instead of the long connections that the original U-Net architecture includes~\cite{zhou2018unet++}. By combining both long and short connections with varying receptive fields, UNet++ has the advantage of capturing features at different levels and integrating them through feature concatenation. This approach elevates its accuracy in performance. Furthermore, the flexible network structure, when coupled with deep supervision, allows for significant reduction in parameter count of deep networks without compromising the acceptable range of accuracy.

SegNet was introduced to address the challenges of semantic image segmentation in autonomous driving and intelligent robotics~\cite{badrinarayanan2016segnet}. It consists of two parts: the Encoder and the Decoder. There exists a symmetric relationship between the Encoder and the Decoder.
The Encoder undertakes feature extraction via convolutions, while pooling enlarges the receptive field, thereby diminishing the image dimensions. Contrarily, the Decoder encompasses deconvolution and upsampling that reconstruct features post image classification, restoring them to their initial dimensions. Finally, the output is passed through softmax, yielding the maximum values for different classes and generating the final segmentation map.

TransUNet utilizes the encoder structure of the transformer in its encoder, allowing for better feature extraction~\cite{chen2021transunet}. The transformer was originally designed for sequence-to-sequence prediction, and its inherent global attention mechanism makes it a viable network structure. However, the transformer is deficient in capturing low-level details, which can result in limited localization capabilities. Due to the limited receptive field, CNNs~\cite{lecun1998gradient,krizhevsky2017imagenet} struggle to effectively utilize global information, although they excel in extracting local details. In contrast, TransUNet combines both approaches, leveraging the advantages of both Transformers and U-Net. The transformer encodes image patches from CNN feature maps as input sequences to extract global features. The decoder upsamples the encoded features and combines them with high-resolution CNN features for precise localization.

Swin-UNet is a pure Transformer architecture similar to U-Net, specifically designed for medical image segmentation, comprising of an Encoder, Bottleneck, Decoder, and Skip Connections~\cite{cao2021swinunet}. The tokenized image blocks are fed into the Transformer-based U-shaped En-Decoder architecture through skip connections to facilitate local and global semantic feature learning. The Swin Transformer blocks are responsible for feature representation learning, while the decoder consists of Swin Transformer blocks and patch expansion layers. The extracted contextual features are fused with multi-scale features from the encoder through skip connections to compensate for spatial information loss caused by downsampling.

\subsection{Methods of detection}
\label{ss:Methods of detection}

Faster R-CNN is the pioneering framework that achieved end-to-end implementation of object detection tasks using deep learning models~\cite{ren2015faster}. It inherits the technical trajectory of both R-CNN and Fast R-CNN~\cite{girshick2015fast} and further introduces the RPN network for efficient and batch-wise generation of region proposals. The algorithm consists of three components: feature extraction by the backbone network, region proposal generation using RPN, and object classification and regression by the RCNN.

SSD (Single Shot MultiBox Detector) enables multi-scale object detection and has achieved superior performance compared to Faster R-CNN on various datasets~\cite{liu2016ssd}. Furthermore, in scenarios where the input image size is small, SSD exhibits higher accuracy compared to other one-stage methods. It introduces a single-stage detector that is more accurate and faster than previous algorithms such as YOLO, without employing RPN and pooling operations. As a result, SSD is simpler than two-stage object detection algorithms

YOLO algorithm is an example of a one-stage object detection algorithm, differing significantly from two-stage object detection algorithms in terms of computational speed. The YOLO series of algorithms divide an image into multiple grids and generate prior bounding boxes based on the anchor mechanism. By generating detection boxes in a single step, this approach greatly enhances the algorithm's prediction speed.

The first improvement in YOLOv3~\cite{redmon2018yolov3} is the replacement of the backbone network with a more effective one, DarkNet53, which extracts relevant features from images to achieve our desired objectives. Compared to DarkNet19 used in YOLOv2~\cite{redmon2017yolo9000}, the new network employs a greater number of convolutions, specifically 53 layers of convolutions. Additionally, it incorporates residual connections from the residual network, thereby enhancing the network's performance.

YOLOv4~\cite{bochkovskiy2020yolov4} improves the Average Precision (AP) and Frames Per Second (FPS) of YOLOv3 by 10\% and 12\% respectively. It incorporates the CSPDarkNet-53 network, which offers excellent trade-offs between speed and accuracy. The core of YOLOv4 is the adoption of the Cross Stage Partial Network structure. Additionally, several improvements are made to the input during training, including Mosaic data augmentation, cmBN (Cross mini-Batch Normalization), and SAT (Self-Adversarial Training).

YOLOv5 demonstrates further improvements in object detection performance compared to YOLOv4. It introduces new techniques such as the PAN structure and adaptive augmentation, resulting in enhanced detection accuracy. YOLOv5 incorporates a novel backbone network architecture called CSPNet~\cite{wang2020cspnet}, which outperforms the DarkNet53 in terms of computational efficiency and accuracy. As a result, YOLOv5 exhibits improvements in both speed and precision.

\subsection{Methods of feature extraction}
\label{ss:Methods of feature extraction}
A total of 1316 texture features were extracted in this study. These features were automatically extracted from the expert-segmented PHE region. They included first-order statistics features~\cite{liew2022large}, shape features~\cite{peng2023pulmonary}, gray-level co-occurrence matrix (GLCM) features~\cite{firdaus2020lung}, gray-level run-length matrix (GLRLM) features, gray-level size-zone matrix (GLSZM) features, gray-level dependence matrix (GLDM) features~\cite{ahmadi2020iris} and neighborhood gray-tone difference matrix (NGTDM) features~\cite{pathak2013texture}. 
A consistency test was performed on the extracted radiomics features using the intra-class correlation coefficient (ICC), and only features with ICC > 0.80 were retained for subsequent modeling.
The purpose of this study is to construct a model based on CT images, using the radiomics features of the PHE around hemorrhage, to predict the risk of HE in the short term in patients with SICH, and to provide better individualized treatment options for the clinic.

\subsubsection{Radiomic features}
\label{sss:Radiomic features}
First-order statistics describe the distribution of voxel intensities within an image region defined by a mask, utilizing commonly used and fundamental measures~\cite{liew2022large}. These features provide information about the distribution and intensity levels of pixels in the image, with a total of 19 features. First-order features encompass several essential attributes of the image. These include Energy, measures the magnitude of voxel values in the image; Entropy, which specifies the uncertainty/randomness of values and quantifies the average information required to encode the image values; Minimum, representing the minimum grayscale value within the region of interest (ROI); Maximum, representing the maximum grayscale intensity within ROI, and others.

Shape features include descriptors for the three-dimensional size and shape of the ROI. These features are independent of the grayscale intensity distribution within the ROI and provide information about the shape, size, and spatial relationships of the objects or regions of interest~\cite{peng2023pulmonary}. Shape features are calculated based on the boundaries or contours of objects. Common shape features include Mesh Volume, Voxel Volume, Surface Area, and others.

Gray Level Co-occurrence Matrix (GLCM)~\cite{firdaus2020lung} is a statistical method used to describe image texture features. It quantifies the spatial relationships between grayscale values of pixels at varying directions and distances within the image~\cite{ozturk2018application}. By constructing a matrix that represents the co-occurrence of gray levels, diverse texture features can be extracted, including joint average, contrast, correlation, and more.

Similarly, the Gray Level Run Length Matrix (GLRLM)~\cite{ozturk2018application} finds application as a statistical method to elucidate the distribution of consecutive gray levels within an image. It computes the frequency of pixel runs with varying lengths for different gray levels, capturing distribution patterns and texture features. Notable GLRLM features encompass short run emphasis, long run emphasis, gray level non-uniformity, and more.

Furthermore, the Gray Level Size Zone Matrix (GLSZM)~\cite{thibault2013advanced} operates as a statistical tool aimed at capturing the distribution of gray levels within regions of diverse sizes in an image. It computes the frequency of gray levels within these differently sized regions, thereby extracting texture features associated with size variation in the image. Notable GLSZM features include small area emphasis, gray level non-uniformity, normalized size-zone non-uniformity, and more~\cite{thibault2013advanced, aerts2014decoding}.

The Gray Level Dependence Matrix (GLDM)~\cite{ahmadi2020iris} takes on the role of a statistical method employed to uncover the interdependence among different gray levels within an image. It quantifies the differences and frequency of pixel-level changes across different gray levels, thereby extracting features related to the dependence of gray levels in the image. Prominent GLDM features comprise small dependence emphasis, large dependence emphasis, and more.

Lastly, the Neighborhood Gray Tone Difference Matrix (NGTDM)~\cite{pathak2013texture} emerges as a statistical approach used to illustrate the variations in gray levels among neighboring pixels within an image. It computes the differences and frequency of gray level changes within diverse pixel neighborhoods, thereby extracting texture features of the image. Noteworthy NGTDM features encompass small dependence emphasis, large dependence emphasis, dependence non-uniformity, and more.

\section{Result}
\label{s:Result}

\subsection{Results of segmentation}
\label{ss:Results of segmentation}
    
\subsubsection{Results of segmentation machine learning methods}
\label{sss:Results of segmentation machine learning methods}

\begin{figure*}[!htbp]
    \centering
    \includegraphics[width=0.9\textwidth]{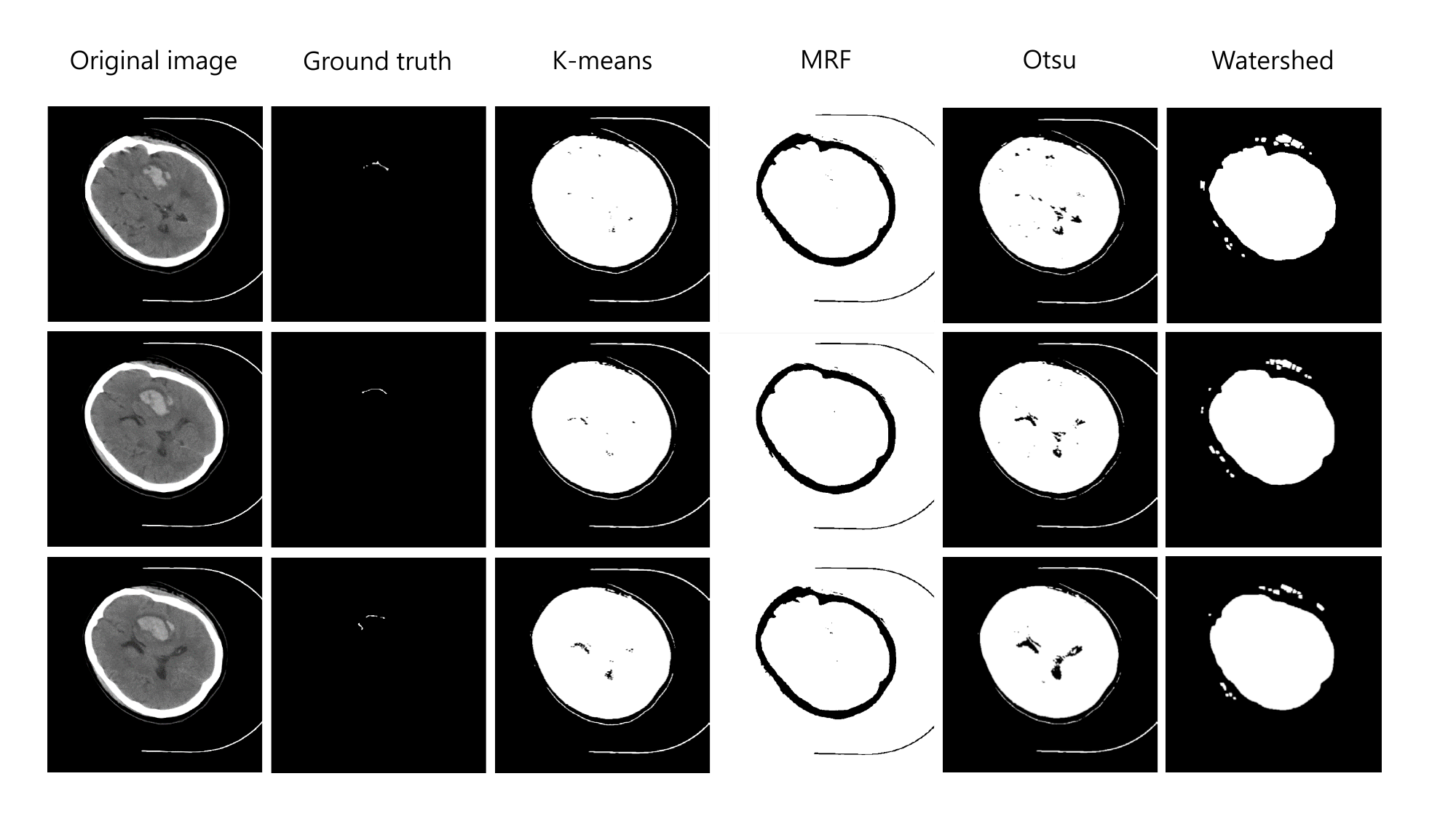}
    \caption{Segmentation results of classical machine learning methods on the PHE-SICH-CT-IDS.}
    \label{fig:machine_seg}
\end{figure*}

Four classical machine learning segmentation methods are experimentally evaluated on the PHE-SICH-CT-IDS, comparing and analyzing the segmentation images and performance under different machine learning methods. The experimental segmentation results are depicted in the Figure~\ref{fig:machine_seg}. The distinct variations in results obtained by applying different classical machine learning segmentation methods demonstrate the effectiveness of PHE-SICH-CT-IDS in evaluating the performance of various segmentation methods. Additionally, the challenging segmentation of PHE highlights the urgent need for a dedicated edema segmentation dataset.


\subsubsection{Results of deep learning methods}
\label{sss:Results of Deep learning methods}

\begin{figure*}[!htbp]
    \centering
    \includegraphics[width=0.9\textwidth]{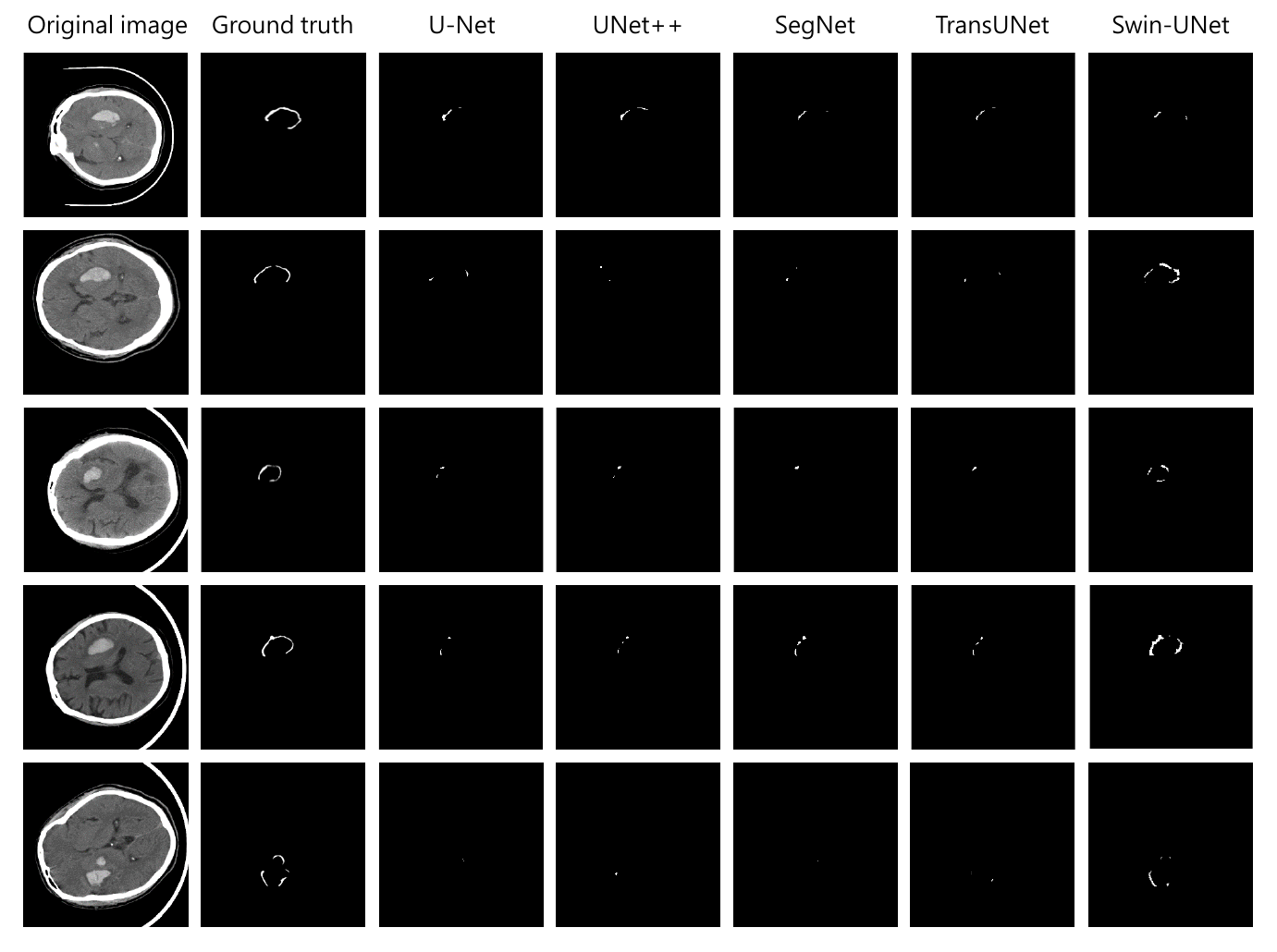}
    \caption{Segmentation results of deep learning methods on the PHE-SICH-CT-IDS.}
    \label{fig:segDeep}
\end{figure*}

The segmentation performance is evaluated on the current dataset using four different deep learning models. In the experiments, each model is trained using a ratio of 4:4:2 for the training, validation, and test sets, respectively. The learning rate is set to 0.000 05, the number of epochs is set to 100, and the batch size is set to 4. Figure~\ref{fig:segDeep} presents the segmentation results obtained using three different models. The evaluation employs five commonly used metrics, including Dice coefficient, Jaccard index, Hausdorff distance, precision, and recall. The evaluation metrics for the segmentation experiments are shown in Table~\ref{s:segDeepTable}.

The experiments were conducted on an NVIDIA GeForce RTX 2080 GPU with 8 GB of memory. In terms of software, the programming was done in Python 3.8, and the PyTorch framework version 1.7.0 was utilized.

\begin{table*}[!htbp]
\label{s:segDeepTable}
\centering
\setlength{\abovecaptionskip}{0pt}
\setlength{\belowcaptionskip}{10pt}
\caption{Evaluation metrics for different segmentation methods based on deep learning.}
\begin{tabular}{cccccc}
\toprule
Methods   & Dice   & Jaccard index & Hausdorff distance & Precision & Recall \\ \midrule
U-Net     & 0.2269 & 0.1488        & 68.2550            & 0.6332    & 0.1582 \\
UNet++    & 0.2178 & 0.1456        & 54.0571            & 0.6086    & 0.1519 \\
SegNet    & 0.1863 & 0.1208        & 48.0345            & 0.5887    & 0.1224 \\
TransUNet & 0.2451 & 0.1552        & 52.4001            & 0.7631    & 0.1591 \\
Swin-UNet & 0.3512 & 0.2285        & 47.8789            & 0.4720    & 0.3321 \\ \bottomrule
\end{tabular}
\end{table*}

\subsection{Results of detection}
\label{ss:Results of detection}

\begin{figure*}[!htbp]
    \centering
    \includegraphics[width=0.9\textwidth]{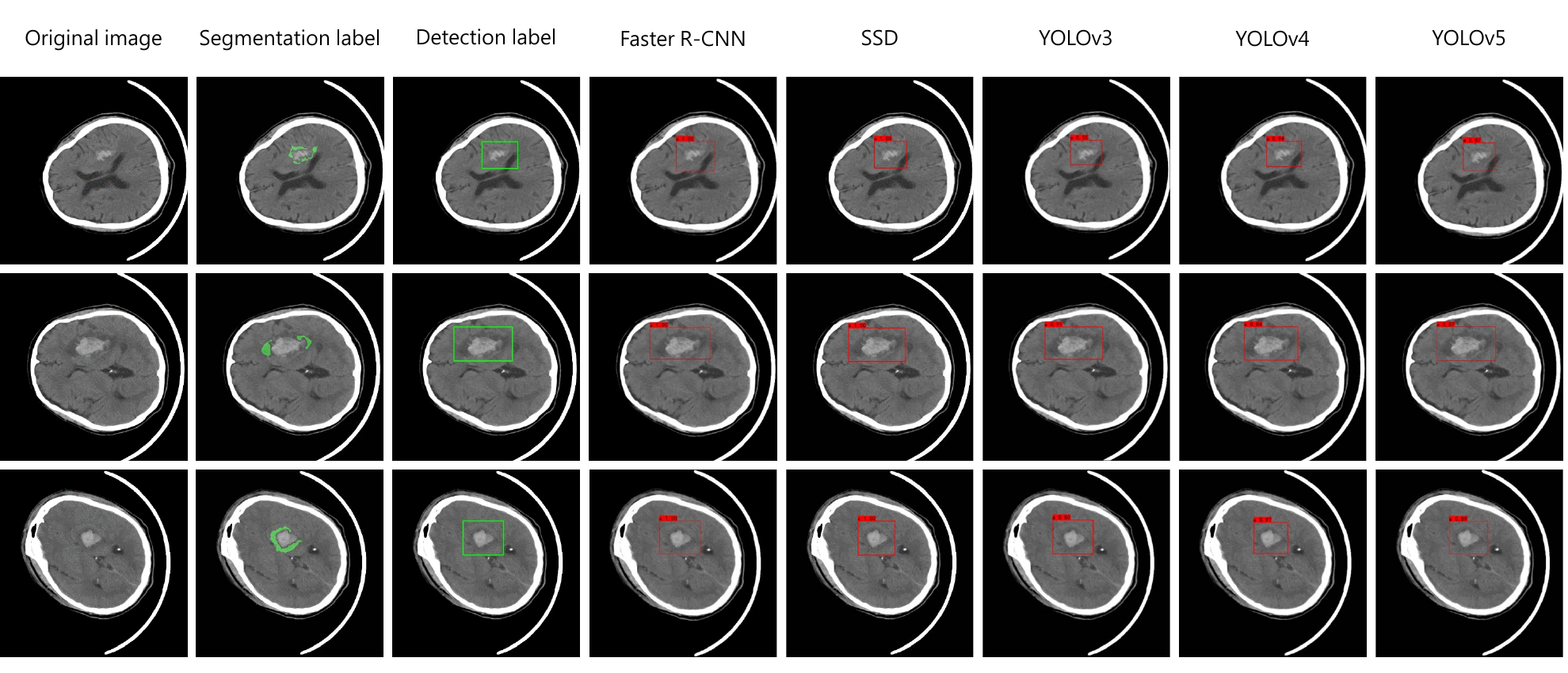}
    \caption{The object detection experiment results on the PHE-SICH-CT-IDS are presented. The first column shows the original CT images, the second column displays the overlay of the segmented edematous regions on the original images, and the third column shows the ground truth bounding boxes of both edematous regions and hemorrhagic lesions. The fourth to eighth columns illustrate the detection results of Faster R-CNN, SSD, YOLOv3, YOLOv4, and YOLOv5, respectively.}
    \label{fig:detectionDeep}
\end{figure*}

In this study, five commonly used object detection models with different architectures are employed to test the feasibility of using the PHE-SICH-CT-IDS for object detection. The targets of detection include not only the hemorrhagic regions but also the edematous zones. The detection results are depicted in Figure~\ref{fig:detectionDeep}. The performance is evaluated using four widely adopted metrics: AP (Average Precision), F1-score, precision, and recall, as shown in Table~\ref{s:detectionEval}. For this experiment, the models are trained for 300 epochs with a learning rate of 0.000 1 and a batch size of 4.

\begin{table}[!htbp]
\label{s:detectionEval}
\centering
\setlength{\abovecaptionskip}{0pt}
\setlength{\belowcaptionskip}{10pt}
\caption{The evaluation results of the object detection experiments on the PHE-SICH-CT-IDS.}
\begin{tabular}{@{}ccccc@{}}
\toprule
Methods      & AP    & F1-score & Precision & Recall \\ \midrule
Faster R-CNN & 96.56 & 0.95     & 91.55     & 96.55  \\
SSD          & 95.44 & 0.94     & 91.89     & 95.77  \\
YOLOv3       & 92.73 & 0.77     & 97.62     & 62.35  \\
YOLOv4       & 94.33 & 0.91     & 94.03     & 88.73  \\
YOLOv5       & 96.37 & 0.96     & 97.10     & 94.37  \\ \bottomrule
\end{tabular}
\end{table}

\subsection{Results of feature extraction}
\label{ss:Results of feature extraction}
The sub-dataset A was randomly partitioned into a training set and a validation set in a 7:3 ratio. The features were normalized using the Z-score algorithm, dimensionality reduction was performed using pearson correlation coefficient (PCC) algorithm. Eight radiomic features were selected using analysis of variance (ANOVA), recursive feature elimination (RFE), and relief algorithms, with their weighted coefficients shown in Table~\ref{s:EightFeatures}. The machine learning model employed logistic regression and incorporated clinical parameters for modeling analysis. The training set achieved an area under curve (AUC) value of 0.835. The validation set yielded an AUC value of 0.745.

\begin{table}[!htbp]
\centering
\setlength{\abovecaptionskip}{0pt}
\setlength{\belowcaptionskip}{10pt}
\caption{The refined features and their corresponding weights after the selection process.}
\label{s:EightFeatures}

\begin{tabular}{@{}cc@{}}
\toprule
Feature                                                   & Weight \\ \midrule
CT\_gradient\_firstorder\_Skewness                        & -0.745 \\
CT\_square\_firstorder\_RobustMeanAbsoluteDeviation       & 1.366  \\
CT\_squareroot\_glcm\_ClusterShade                        & -0.601 \\
CT\_squareroot\_gldm\_LargeDependenceLowGrayLevelEmphasis & 0.651  \\
CT\_squareroot\_glszm\_LowGrayLevelZoneEmphasis           & 0.092  \\
CT\_wavelet-HLH\_glszm\_SmallAreaHighGrayLevelEmphasis    & 0.229  \\
CT\_wavelet-HLL\_glcm\_Correlation                        & -0.707 \\
CT\_wavelet-LLL\_firstorder\_Maximum                      & -1.069 \\ \bottomrule
\end{tabular}
\end{table}


\section{Discussion}
\label{s:Discussion}
\subsection{Segmentation results discussion}
\label{ss:Segmentation results discussion}
According to Figure~\ref{fig:machine_seg}, the PHE-SICH-CT-IDS exhibits noticeable variations in segmentation results when different classic machine learning segmentation methods are employed. However, it is challenging to accurately segment the PHE in CT images due to several reasons. Firstly, the edematous region associated with cerebral hemorrhage exhibits similarity in morphology and density with surrounding tissues, making it difficult to achieve clear segmentation from the surrounding structures. This similarity can result in errors in machine learning models, particularly in boundary regions or in edematous areas with complex shapes. Secondly, the edematous region displays significant variability across different CT images, including differences in size, shape, and location. This variability poses a challenge in building a universal model capable of accommodating different patients, scanners, and scanning parameters, thereby limiting the model's generalizability. As a result, classical machine learning methods face significant difficulties in accurately segmenting the edematous region. 
However, in this study, distinct segmentation results were achieved for CT images. There are noticeable differences among different classic machine learning segmentation methods. Hence, the PHE-SICH-CT-IDS proves to be effective in assessing the segmentation performance of different classic machine learning segmentation methods.

Overall, deep learning methods outperform classical machine learning methods. Comparing the results in Figure~\ref{fig:machine_seg} and Figure~\ref{fig:segDeep}, it is evident that deep learning methods accurately locate and segment the objects. In the PHE-SICH-CT-IDS, the Swin-UNet achieves the highest Dice coefficient, Jaccard index, and recall, reaching 0.35, 0.23, and 0.33, respectively. The TransUNet achieves the highest precision at 0.76. It can be observed that the two models combining the Transformer~\cite{vaswani2017attention} with the U-Net architecture perform the best. However, these models have larger sizes and longer training times. Among them, SegNet has the shortest training time and the smallest model size, but it also has the poorest performance among the five models. U-Net and UNet++ exhibit comparable performance. However, they demonstrate significant dissimilarity in terms of Hausdorff distance.

\subsection{Detection results discussion}
\label{ss:Detection results discussion}

Based on the observation in Figure~\ref{fig:detectionDeep}, the object detection algorithms demonstrate excellent performance on the PHE-SICH-CT-IDS, accurately detecting almost all targets, including hemorrhagic regions and PHE zones. Table~\ref{s:detectionEval} presents significant variations in various evaluation metrics among the different algorithms. Notably, YOLOv5 exhibits the best overall performance in object detection, achieving the highest precision of around 97$\%$. However, YOLOv3 shows the lowest F1-score and recall, at 0.77 and 62.35$\%$, respectively. Additionally, SSD exhibits the fastest detection speed among the tested methods. Therefore, the PHE-SICH-CT-IDS proves to be effectively utilized for image object detection.

\subsection{Feature extraction results discussion}
\label{ss:Feature extraction results discussion}
This study demonstrates the value of a combined model constructed using texture features extracted from CT images for prognostic predictions in patients with SICH. By applying this combined model, frontline clinicians can self-assess the risk of hematoma expansion in SICH patients and develop personalized treatment strategies. Based on these findings, it is concluded that PHE-SICH-CT-IDS can be utilized to evaluate image features.


\section{Conclusion and futures works}
\label{s:Conclusion and futures works}

This study developed the first open-source CT image dataset, named PHE-SICH-CT-IDS, specifically designed for segmenting PHE in patients with spontaneous basal ganglia intracerebral hemorrhage. In addition to the segmentation of PHE, the dataset provides functionality for hemorrhage detection and radiomic feature extraction. PHE-SICH-CT-IDS consists of three sub-datasets with different data types, allowing users to select the appropriate dataset based on their specific needs. To evaluate the segmentation performance, we employ commonly used machine learning methods as well as five deep learning segmentation methods, including U-Net and Swin-UNet. The experimental results demonstrated that PHE-SICH-CT-IDS can discern between different segmentation methods, supported by multiple evaluation metrics. Furthermore, five object detection methods have been used during the experiments such as Faster R-CNN and YOLOv5, which accurately identify the hemorrhage region and the surrounding edema. Thus, PHE-SICH-CT-IDS holds promise for effective image-based object detection. Additionally, we extract a wide range of radiomic features and construct a joint model combining logistic regression and clinical parameters. The classification performance of this model was significantly improved, providing valuable assistance to frontline clinicians in assessing the HE risk in patients with SICH.

In the forthcoming future, the dataset will undergo expansion to encompass a broader scope of medical scenarios. Concurrently, efforts will focus on refining and advancing the segmentation methods to provide a more sophisticated and robust resource for the medical community. This ongoing pursuit of excellence aims to empower medical professionals with cutting-edge tools and foster groundbreaking advancements in the realm of medical research.

\section*{Acknowledgments}
This work is supported by the ``National Natural Science Foundation of China'' (No. 82220108007). We thank B.A. Qi Qiu, from Foreign Studies College in Northeastern University, China, for her professional English proofreading in this paper. We also thank Miss. Zixian Li and Mr. Guoxian Li for their important discussion in this work.

\section*{Declaration of Competing Interest}
The authors declare that they have no conflict of interest in this paper.

\section*{Data Cvailability Statement}

The datasets featured in this study are available in online repositories. 
The repository name and accession number(s) can be found below: \url{https://figshare.com/articles/dataset/PHE-SICH-CT-IDS/23957937}.

\bibliographystyle{IEEEtran}
\bibliography{references}

\begin{thebibliography}{10}
\providecommand{\url}[1]{#1}
\csname url@samestyle\endcsname
\providecommand{\newblock}{\relax}
\providecommand{\bibinfo}[2]{#2}
\providecommand{\BIBentrySTDinterwordspacing}{\spaceskip=0pt\relax}
\providecommand{\BIBentryALTinterwordstretchfactor}{4}
\providecommand{\BIBentryALTinterwordspacing}{\spaceskip=\fontdimen2\font plus
\BIBentryALTinterwordstretchfactor\fontdimen3\font minus
  \fontdimen4\font\relax}
\providecommand{\BIBforeignlanguage}[2]{{%
\expandafter\ifx\csname l@#1\endcsname\relax
\typeout{** WARNING: IEEEtran.bst: No hyphenation pattern has been}%
\typeout{** loaded for the language `#1'. Using the pattern for}%
\typeout{** the default language instead.}%
\else
\language=\csname l@#1\endcsname
\fi
#2}}
\providecommand{\BIBdecl}{\relax}
\BIBdecl

\bibitem{Example2020}
E.~Author, ``My article,'' 2020.

\bibitem{Li-2016-CBMIA}
C.~Li, \emph{{Content-based Microscopic Image Analysis}}.\hskip 1em plus 0.5em
  minus 0.4em\relax Germany: Logos Verlag Berlin GmbH, 2016.

\bibitem{We-1955-SOB}
T.~We, ``{Section of Biology: the Cytoanalyzer-an Example of Physics in Medical
  Research},'' \emph{Transactions of the New York Academy of Sciences},
  vol.~17, no.~3, pp. 250--256, 1955.

\bibitem{Yang-2005-CST}
F.~Yang, M.~A. Mackey, and F.~Lanzini, ``{Cell Segmentation, Tracking, and
  Mitosis Detection Using Temporal Context},'' in \emph{Medical Image Computing
  and Computer-assisted Intervention}, 2005, pp. 302--309.

\bibitem{Sobel-2014-HAD}
I.~Sobel, ``{History and definition of the Sobel operator},'' 2014.

\bibitem{Yang-2008-ASO}
M.~Yang, K.~Kpalma, and J.~Ronsin, ``{A Survey of Shape Feature Extraction
  Techniques},'' in \emph{{Pattern Recognition}}, P.~Yin, Ed.\hskip 1em plus
  0.5em minus 0.4em\relax {IN-TECH}, 2008, pp. 43--90.

\bibitem{2012Intracerebral}
R.~F. Keep, Y.~Hua, and G.~Xi, ``Intracerebral haemorrhage: mechanisms of
  injury and therapeutic targets,'' \emph{The Lancet Neurology}, vol.~11,
  no.~8, pp. 720--731, 2012.

\bibitem{2019Heart}
E.~J. Benjamin, P.~Muntner, and A.~Alonso, ``Heart disease and stroke
  statistics-2019 update: A report from the american heart association,''
  \emph{Circulation}, vol. 139, no.~10, p. e56–528, 2019.

\bibitem{2010Advances}
O.~Adeoye and J.~P. Broderick, ``Advances in the management of intracerebral
  hemorrhage,'' \emph{Nature Reviews Neurology}, vol.~6, no.~11, pp. 593--601,
  2010.

\bibitem{2010Incidence}
C.~Asch, M.~J. Luitse, G.~J. Rinkel, I.~Tweel, and C.~J. Klijn, ``Incidence,
  case fatality, and functional outcome of intracerebral haemorrhage over time,
  according to age, sex, and ethnic origin: a systematic review and
  meta-analysis.'' \emph{Lancet Neurology}, vol.~9, no.~2, pp. 167--176, 2010.

\bibitem{2015Emergency}
E.~C. Jauch, J.~A. Pineda, and J.~C. Hemphill, ``Emergency neurological life
  support: Intracerebral hemorrhage,'' \emph{Neurocritical Care}, vol.~23, no.
  2 Supplement, pp. 83--93, 2015.

\bibitem{2019Recent}
R.~Garg and J.~Biller, ``Recent advances in spontaneous intracerebral
  hemorrhage,'' \emph{F1000 Research}, vol.~8, 2019.

\bibitem{2017Classification}
X.~W. Gao, R.~Hui, and Z.~Tian, ``Classification of ct brain images based on
  deep learning networks,'' \emph{Computer Methods & Programs in Biomedicine},
  vol. 138, pp. 49--56, 2017.

\bibitem{C2014Automated}
C.~R.~G. a, G.~W.~H. A, and D.~M.~A. B, ``Automated delineation of stroke
  lesions using brain ct images,'' \emph{NeuroImage: Clinical}, vol.~4, no.~C,
  pp. 540--548, 2014.

\bibitem{2011Defining}
D.~Dowlatshahi, A.~M. Demchuk, M.~L. Flaherty, M.~Ali, P.~L. Lyden, and E.~E.
  Smith, ``Defining hematoma expansion in intracerebral hemorrhage:
  Relationship with patient outcomes,'' \emph{Neurology}, vol.~76, no.~14, pp.
  1238--1244, 2011.

\bibitem{ronneberger2015unet}
O.~Ronneberger, P.~Fischer, and T.~Brox, ``U-net: Convolutional networks for
  biomedical image segmentation,'' 2015.

\bibitem{chilamkurthy2018development}
S.~Chilamkurthy, R.~Ghosh, S.~Tanamala, M.~Biviji, N.~G. Campeau, V.~K.
  Venugopal, V.~Mahajan, P.~Rao, and P.~Warier, ``Development and validation of
  deep learning algorithms for detection of critical findings in head ct
  scans,'' 2018.

\bibitem{chilamkurthy2018deep}
S.~Chilamkurthy, R.~Ghosh, S.~Tanamala, M.~Biviji, and Campeau, ``Deep learning
  algorithms for detection of critical findings in head ct scans: a
  retrospective study,'' \emph{The Lancet}, vol. 392, no. 10162, pp.
  2388--2396, 2018.

\bibitem{lee2019explainable}
H.~Lee, S.~Yune, M.~Mansouri, M.~Kim, S.~H. Tajmir, C.~E. Guerrier, S.~A.
  Ebert, S.~R. Pomerantz, J.~M. Romero, S.~Kamalian \emph{et~al.}, ``An
  explainable deep-learning algorithm for the detection of acute intracranial
  haemorrhage from small datasets,'' \emph{Nature biomedical engineering},
  vol.~3, no.~3, pp. 173--182, 2019.

\bibitem{MD2020Intracranial}
M.~D. Hssayeni, M.~S. Croock, A.~D. Salman, H.~F. Al-khafaji, Z.~A. Yahya, and
  B.~Ghoraani, ``Intracranial hemorrhage segmentation using a deep
  convolutional model,'' \emph{Data}, vol.~5, no.~1, p.~14, 2020.

\bibitem{hssayeni2020computed}
M.~Hssayeni, M.~Croock, A.~Salman, H.~Al-khafaji, Z.~Yahya, and B.~Ghoraani,
  ``Computed tomography images for intracranial hemorrhage detection and
  segmentation,'' \emph{Intracranial Hemorrhage Segmentation Using A Deep
  Convolutional Model. Data}, vol.~5, no.~1, p.~14, 2020.

\bibitem{chen2022defining}
Y.~Chen, C.~Qin, J.~Chang, Y.~Liu, Q.~Zhang, Z.~Ye, Z.~Li, F.~Tian, W.~Ma,
  J.~Wei \emph{et~al.}, ``Defining delayed perihematomal edema expansion in
  intracerebral hemorrhage: segmentation, time course, risk factors and
  clinical outcome,'' \emph{Frontiers in Immunology}, vol.~13, 2022.

\bibitem{kuang2022uncertainty}
Z.~Kuang, Z.~Yan, L.~Yu, X.~Deng, Y.~Hua, and S.~Li, ``Uncertainty-aware deep
  learning with cross-task supervision for phe segmentation on ct images,''
  \emph{IEEE Journal of Biomedical and Health Informatics}, vol.~26, no.~6, pp.
  2615--2626, 2022.

\bibitem{flanders2020construction}
A.~E. Flanders, L.~M. Prevedello, G.~Shih, S.~S. Halabi, J.~Kalpathy-Cramer,
  R.~Ball, J.~T. Mongan, A.~Stein, F.~C. Kitamura, M.~P. Lungren \emph{et~al.},
  ``Construction of a machine learning dataset through collaboration: the rsna
  2019 brain ct hemorrhage challenge,'' \emph{Radiology: Artificial
  Intelligence}, vol.~2, no.~3, p. e190211, 2020.

\bibitem{liew2022large}
S.-L. Liew, B.~P. Lo, M.~R. Donnelly, A.~Zavaliangos-Petropulu, J.~N. Jeong,
  G.~Barisano, A.~Hutton, J.~P. Simon, J.~M. Juliano, A.~Suri \emph{et~al.},
  ``A large, curated, open-source stroke neuroimaging dataset to improve lesion
  segmentation algorithms,'' \emph{Scientific data}, vol.~9, no.~1, p. 320,
  2022.

\bibitem{aggarwal2012first}
N.~Aggarwal and R.~Agrawal, ``First and second order statistics features for
  classification of magnetic resonance brain images,'' 2012.

\bibitem{peng2023pulmonary}
Y.~Peng, P.~Luan, H.~Tu, X.~Li, and P.~Zhou, ``Pulmonary fissure segmentation
  in ct images based on odos filter and shape features,'' \emph{Multimedia
  Tools and Applications}, pp. 1--22, 2023.

\bibitem{firdaus2020lung}
Q.~Firdaus, R.~Sigit, T.~Harsono, and A.~Anwar, ``Lung cancer detection based
  on ct-scan images with detection features using gray level co-occurrence
  matrix (glcm) and support vector machine (svm) methods,'' in \emph{2020
  International Electronics Symposium (IES)}.\hskip 1em plus 0.5em minus
  0.4em\relax IEEE, 2020, pp. 643--648.

\bibitem{ahmadi2020iris}
N.~Ahmadi and G.~Akbarizadeh, ``Iris tissue recognition based on gldm feature
  extraction and hybrid mlpnn-ica classifier,'' \emph{Neural Computing and
  Applications}, vol.~32, pp. 2267--2281, 2020.

\bibitem{pathak2013texture}
B.~Pathak and D.~Barooah, ``Texture analysis based on the gray-level
  co-occurrence matrix considering possible orientations,'' \emph{International
  Journal of Advanced Research in Electrical, Electronics and Instrumentation
  Engineering}, vol.~2, no.~9, pp. 4206--4212, 2013.

\bibitem{ozturk2018application}
{\c{S}}.~{\"O}zt{\"u}rk and B.~Akdemir, ``Application of feature extraction and
  classification methods for histopathological image using glcm, lbp, lbglcm,
  glrlm and sfta,'' \emph{Procedia computer science}, vol. 132, pp. 40--46,
  2018.

\bibitem{thibault2013advanced}
G.~Thibault, J.~Angulo, and F.~Meyer, ``Advanced statistical matrices for
  texture characterization: application to cell classification,'' \emph{IEEE
  Transactions on Biomedical Engineering}, vol.~61, no.~3, pp. 630--637, 2013.

\bibitem{aerts2014decoding}
H.~J. Aerts, E.~R. Velazquez, R.~T. Leijenaar, C.~Parmar, P.~Grossmann,
  S.~Carvalho, J.~Bussink, R.~Monshouwer, B.~Haibe-Kains, D.~Rietveld
  \emph{et~al.}, ``Decoding tumour phenotype by noninvasive imaging using a
  quantitative radiomics approach,'' \emph{Nature communications}, vol.~5,
  no.~1, p. 4006, 2014.

\bibitem{2014Very}
K.~Simonyan and A.~Zisserman, ``Very deep convolutional networks for
  large-scale image recognition,'' \emph{Computer Science}, 2014.

\bibitem{zhou2018unet++}
Z.~Zhou, M.~M. Rahman~Siddiquee, N.~Tajbakhsh, and J.~Liang, ``Unet++: A nested
  u-net architecture for medical image segmentation,'' in \emph{Deep Learning
  in Medical Image Analysis and Multimodal Learning for Clinical Decision
  Support: 4th International Workshop, DLMIA 2018, and 8th International
  Workshop, ML-CDS 2018, Held in Conjunction with MICCAI 2018, Granada, Spain,
  September 20, 2018, Proceedings 4}.\hskip 1em plus 0.5em minus 0.4em\relax
  Springer, 2018, pp. 3--11.

\bibitem{badrinarayanan2016segnet}
V.~Badrinarayanan, A.~Kendall, and R.~Cipolla, ``Segnet: A deep convolutional
  encoder-decoder architecture for image segmentation,'' 2016.

\bibitem{chen2021transunet}
J.~Chen, Y.~Lu, Q.~Yu, X.~Luo, E.~Adeli, Y.~Wang, L.~Lu, A.~L. Yuille, and
  Y.~Zhou, ``Transunet: Transformers make strong encoders for medical image
  segmentation,'' 2021.

\bibitem{cao2021swinunet}
H.~Cao, Y.~Wang, J.~Chen, D.~Jiang, X.~Zhang, Q.~Tian, and M.~Wang,
  ``Swin-unet: Unet-like pure transformer for medical image segmentation,'' in
  \emph{European conference on computer vision}.\hskip 1em plus 0.5em minus
  0.4em\relax Springer, 2022, pp. 205--218.

\bibitem{lecun1998gradient}
Y.~LeCun, L.~Bottou, Y.~Bengio, and P.~Haffner, ``Gradient-based learning
  applied to document recognition,'' \emph{Proceedings of the IEEE}, vol.~86,
  no.~11, pp. 2278--2324, 1998.

\bibitem{krizhevsky2017imagenet}
A.~Krizhevsky, I.~Sutskever, and G.~E. Hinton, ``Imagenet classification with
  deep convolutional neural networks,'' \emph{Communications of the ACM},
  vol.~60, no.~6, pp. 84--90, 2017.

\bibitem{ren2015faster}
S.~Ren, K.~He, R.~Girshick, and J.~Sun, ``Faster r-cnn: Towards real-time
  object detection with region proposal networks,'' \emph{Advances in neural
  information processing systems}, vol.~28, 2015.

\bibitem{liu2016ssd}
W.~Liu, D.~Anguelov, D.~Erhan, C.~Szegedy, S.~Reed, C.-Y. Fu, and A.~C. Berg,
  ``Ssd: Single shot multibox detector,'' in \emph{Computer Vision--ECCV 2016:
  14th European Conference, Amsterdam, The Netherlands, October 11--14, 2016,
  Proceedings, Part I 14}.\hskip 1em plus 0.5em minus 0.4em\relax Springer,
  2016, pp. 21--37.

\bibitem{redmon2017yolo9000}
J.~Redmon and A.~Farhadi, ``Yolo9000: better, faster, stronger,'' in
  \emph{Proceedings of the IEEE conference on computer vision and pattern
  recognition}, 2017, pp. 7263--7271.

\bibitem{redmon2018yolov3}
------, ``Yolov3: An incremental improvement,'' \emph{arXiv preprint
  arXiv:1804.02767}, 2018.

\bibitem{bochkovskiy2020yolov4}
A.~Bochkovskiy, C.-Y. Wang, and H.-Y.~M. Liao, ``Yolov4: Optimal speed and
  accuracy of object detection,'' \emph{arXiv preprint arXiv:2004.10934}, 2020.

\bibitem{girshick2015fast}
R.~Girshick, ``Fast r-cnn,'' in \emph{Proceedings of the IEEE international
  conference on computer vision}, 2015, pp. 1440--1448.

\bibitem{wang2020cspnet}
C.-Y. Wang, H.-Y.~M. Liao, Y.-H. Wu, P.-Y. Chen, J.-W. Hsieh, and I.-H. Yeh,
  ``Cspnet: A new backbone that can enhance learning capability of cnn,'' in
  \emph{Proceedings of the IEEE/CVF conference on computer vision and pattern
  recognition workshops}, 2020, pp. 390--391.

\bibitem{2015Image}
N.~Dhanachandra, K.~Manglem, and Y.~J. Chanu, ``Image segmentation using
  k-means clustering algorithm and subtractive clustering algorithm,''
  \emph{Procedia Computer Science}, vol.~54, pp. 764--771, 2015.

\bibitem{2004Unsupervised}
H.~Deng and D.~A. Clausi, ``Unsupervised image segmentation using a simple mrf
  model with a new implementation scheme,'' 2004, pp. 2323--2335.

\bibitem{liu2009otsu}
D.~Liu and J.~Yu, ``Otsu method and k-means,'' in \emph{2009 Ninth
  International conference on hybrid intelligent systems}, vol.~1.\hskip 1em
  plus 0.5em minus 0.4em\relax IEEE, 2009, pp. 344--349.

\bibitem{shojaii2005automatic}
R.~Shojaii, J.~Alirezaie, and P.~Babyn, ``Automatic lung segmentation in ct
  images using watershed transform,'' in \emph{IEEE international conference on
  image processing 2005}, vol.~2.\hskip 1em plus 0.5em minus 0.4em\relax IEEE,
  2005, pp. II--1270.

\bibitem{vaswani2017attention}
A.~Vaswani, N.~Shazeer, N.~Parmar, J.~Uszkoreit, L.~Jones, A.~N. Gomez, L.~u.
  Kaiser, and I.~Polosukhin, ``Attention is all you need,'' in \emph{Advances
  in Neural Information Processing Systems}, I.~Guyon, U.~V. Luxburg,
  S.~Bengio, H.~Wallach, R.~Fergus, S.~Vishwanathan, and R.~Garnett, Eds.,
  vol.~30.\hskip 1em plus 0.5em minus 0.4em\relax Curran Associates, Inc.,
  2017.

\end{thebibliography}

\end{document}